# Spintronic Physical Reservoir for Autonomous Prediction and Long-Term Household Energy Load Forecasting

**Walid Al Misba[1], Harindra S. Mavikumbure[2], Md Mahadi Rajib[1], Daniel L. Marino[2], Victor Cobilean[2], Milos Manic[2], Fellow, IEEE, and Jayasimha Atulasimha[1,3], Senior Member, IEEE**

[1]Mechanical and Nuclear Engineering Department, Virginia Commonwealth University, Richmond, VA 23284 USA
[2] Computer Science Department, Virginia Commonwealth University, Richmond, VA 23284 USA
[3]Electrical and Computer Engineering Department, Virginia Commonwealth University, Richmond, VA 23284 USA

Corresponding author: (e-mail:) misbawa@vcu.edu, jatulasimha@vcu.edu

This work was supported in part by the Virginia Commonwealth Cyber Initiative (CCI) CCI Cybersecurity Research Collaboration Grant and the Central Virginia Node (CVN) of the Commonwealth Cyber Initiative (CCI) research award number VV-1Q23-008.

**ABSTRACT** With the growing use of artificial neural networks (ANNs) in temporal data processing tasks, the cost of training for complex ANNs is an increasing concern. Physical reservoir computing (RC), a variation of recurrent neural networks (RNNs), obviates the need for most data intensive matrix vector multiplication in the recurrent layer by evolving the RC's internal states with the inherent nonlinear dynamics and short-term memory. In this study, we show that magnetic skyrmion confined in a fixed geometry forming the soft layer of a magnetic tunnel junction (MTJ) can work as a RC and perform autonomous long-term prediction of temporal data. Our proposed skyrmion reservoir allows for manipulation of spin dynamics with ultra energy efficient voltage controlled magnetic anisotropy (VCMA) modulation method. Furthermore, the boundary effect on the skyrmion from the geometric edges provides necessary consistency property of the reservoir. We employ our proposed reservoir for the modeling and prediction of the chaotic time series such as Mackey-Glass and dynamic time-series data, such as household building energy loads. For autonomous run, the predicted output is fed to the input of the reservoir. By comparing our spintronic physical RC approach with energy load forecasting algorithms, such as long short-term memory (LSTM) and RNNs, we conclude that the proposed framework presents good performance in achieving high predictions accuracy, while also requiring low memory and energy both of which are at a premium in hardware resource and power constrained edge applications. Further, the proposed approach is shown to require very small training datasets and at the same time being at least 16× energy efficient compared to the sequence to sequence LSTM for accurate household load predictions. Higher endurance, fast processing and well-established technology (i.e., MTJ) to integrate with CMOS technology makes such spintronic reservoir attractive over other emergent memory device-based reservoirs.

## I. INTRODUCTION

Recurrent neural networks (RNNs) [1,2] are shown to be more suitable in temporal data processing tasks than the traditional feedforward neural networks (FNNs) because of the recurrent connections among constituent neurons. However, RNNs often suffer from vanishing and exploding gradients problem due to the long-term dependencies that could arise in the recurrent layers. To circumvent these issues variations of RNN is proposed, i.e., LSTMs [3] and reservoir computing (RC) [4,5]. In RC, the reservoir consists of an RNN which maps temporal inputs to higher dimensional features due to the short-term memory property that exists in the reservoir and a read-out layer that analyzes the features stored as reservoir states. The RNN connections are fixed and only the read-out layer is trained [4]. Thus, the training can be performed with simple learning rules such as linear regression which makes RC much simpler to implement with low training cost. Recently software-based RC systems have been shown to achieve state-of-the-art performance in speech recognition tasks [6] and superior performances in forecasting tasks, such as prediction of financial systems [7], water inflow [8], and chaotic system prediction [9].

Since the essence of RC is to employ non-linearity to transform input to high dimensional space, any physical dynamic non-linear system can work as a reservoir. For a typical RNN implemented on hardware, the required training is performed for all layers of the neural network [10]. This can be implemented on neuromorphic chips [11]. However, in RC the inference is performed using physical phenomena, and only linear regression is used to train weights between select physical reservoir states and the output. This makes information processing much faster and involves low training cost. These features make physical systems the preferred candidate for hardware implementations of RC. Towards this end, the choice of physical reservoir remains explorative, and researchers investigated electronic [12, 13], photonic [14, 15], memristive [16, 17], spintronic [18-22] reservoir and so on. The spintronic reservoir is most attractive due to its significantly higher endurance cycle and faster information processing. For instance, spintronic computational memory has an endurance over ~ $10^{15}$ cycle and write speed of ~ 1-10 ns [23]. Whereas the phase-change memory based memristor device exhibits an endurance of ~ $10^9$ cycle and has a write speed ~ 100 ns [24]. Also, the read (i.e., magnetoresistance) and write techniques (i.e., spin torques) of spintronic devices and associated integration with CMOS technologies are well established due to their historical use as magnetic hard drives, sensors and magnetic random-access memory devices [25]. In spintronic reservoir only one small-scale non-linear node (i.e., nanomagnets) can essentially capture non-linear dynamics. In comparison, electronic reservoirs (i.e., Mackey-Glass nonlinear circuit element with delayed feedback) [12], use a number of active (such as op-amps) and passive elements (such as resistors) for such capability which could cause prohibitive energy and memory footprint. While the photonic reservoir allows for faster processing, compact design with short time delays requires extremely fast input and output processing, a significant drawback for practical implementation [5]. Moreover, short range exchange interaction and long-range dipolar interaction in spintronic systems provides the opportunity to couple magnetic nodes without any physical interconnections and allow more complex coupling otherwise absent in above-mentioned alternative reservoirs.

Various devices concepts are proposed for spintronic reservoir including nanoscale magnetic structures such as spin torque nano-oscillators [18], planner ensemble of nanomagnets interacting in dipole coupled [26,27] and spin wave mediated [20,28] systems and artificial spin ice (ASI) [29]. Chiral magnetic textures such as domain walls (DWs) [21,30], skyrmions [22,31] and skyrmion lattice [19,32] are also proposed. Individually accessing the dense arrays of ASI and planar nanomagnets [26] with addressable MTJs remains a fabrication challenge with modern technology. Furthermore, recent experiments with ferromagnetic resonance with ASI [29] and interconnected ring arrays with magnetic DWs [30] require external magnetic field for dynamic interaction, which is energy prohibitive. Moreover, low loss propagation of spin wave requires higher quality crystal growth (Yttrium Iron Garnet) and fabrication of nano-antennas to excite and detect spin waves which are prone to Ohmic losses [33]. While the spin-torque nano-oscillator based MTJs [18] shows excellent performance with a greater promise of low area footprint and energy cost, the fast-oscillatory signal cannot be used directly for postprocessing (microwave diode is used to capture amplitude variation).

In contrast, chiral spin texture, skyrmion, confined in a fixed geometry working as a RC allows for external magnetic field free control and direct processing of the generated magnetoresistance signal. Moreover, skyrmion shows higher mobility and has low pinning potential than other chiral structures such as DWs, thus can be excited and translated with extremely low amplitude excitation [19]. In addition, confined skyrmion textures stacked within an MTJ device allows for ultra-energy efficient VCMA modulation [34-39]. Furthermore, confinement of skyrmion provides necessary repulsion from boundary which ensures an essential consistency property to the reservoir [40], where the skyrmion needs to be relaxed to the same energy minima upon withdrawing the input excitation.

Although physical RCs have been shown to implement prediction tasks, most of the works attempt to perform one-step ahead prediction. Long term prediction is important for real-world data as future evolution can facilitate more informed decision and customized policy making. However, multi-step prediction itself is challenging due to the non-linear nature of most real-world data and typically inaccurate predictions of the immediate future accumulate very fast and cause divergence in the future predictions over longer times. Authors in [41] shown multi-step prediction for high spatial dimension data using spatiotemporal transformation and encoder-decoder like reservoir. However, this approach can fall short for univariate data such as individual household power consumption prediction. In household load forecasting tasks, usually the real time power value is readily available from wattmeter, however, the voltage, current values and other parameters are unknown. In [42], autonomous multi-step prediction has been shown for chaotic Mackey-Glass (MG) series by feeding delayed output to the reservoir. However, the reservoir response is transformed using a non-linear function which could be costly, and also diminishes the benefits obtained from linear reservoir operation. Recently, delayed inputs [43] and polynomial transformation of the delayed inputs [44] are used to improve the prediction performance of the reservoir, however, these works attempt to solve the optimized one-step ahead prediction. In this study, we have shown multi-step autonomous prediction using spintronic magnetic skyrmion based RC system. For autonomous prediction, the predicted output is fed directly to the input. To adequately extract the non-linearity that arises in the reservoir dynamics, we include several previous states of the reservoir during the training. This obviates the need of performing any non-linear (i.e. trigonometric, polynomial) transformation of the reservoir states.

For long-term prediction our skyrmion based RC employs the virtual node concept originally proposed in [45] and shown in Fig. 1. Instead of fabricating large number of reservoir nodes to increase the reservoir dimensionality, a single physical nonlinear node subjected to delayed feedback acts as a chain of virtual nodes. At first, we have shown long term prediction for chaotic MG time series since it has been frequently used for benchmarking forecasting tasks. Moreover, the long-term prediction can diverge more quickly due to the sensitivity of the chaotic systems to the error. Next, we have shown individual household power demand forecasting, which is an active research area. According to a recent study, the amount of energy wasted in a commercial building, can reach up to 40% if energy consumption is not properly maintained [46]. With energy management system (EMS), a commercial building can save up to 25.6% of its total energy consumption [47]. EMS in a building requires accurate load forecasting to maintain stability, improve performance, and detect abnormal system behavior. However, accurate long-term forecasting especially in a single household building is very challenging due to the volatile and univariate nature of the household power consumption data. Traditional statistical approaches such as auto-regressive moving average (ARIMA) [48], time-series statistical model [49] suffer from low prediction accuracy due to the parameters assumed in the model and the complexity of the systems. Machine learning based models such as RNN [50] and LSTM [51] offer more flexibility in this regard as they do not depend on the parameters of the system. Instead, they are driven by the observed past and present data, however, with significant training cost. RC can perform the prediction tasks with much more efficiency due to its low training cost and thus is very much suitable to be implemented in edge computing platforms which are equipped with low power devices.

We use three decoupled and patterned skyrmion devices as our reservoir where the temporal correlation of the inputs is captured by the inherent short-term memory of the breathing skyrmions. Upon excitation, the skrymions undergoes oscillations and the skyrmion states are read at a regular interval and processed with linear regression for the prediction. Autonomous prediction up to 30-time steps for the MG time series and up to 23 hours (equivalent to 23-time steps as only hourly demand is typically required) for the household power prediction have been demonstrated by using the predicted output as the input for the next time step prediction.

The rest of the paper is organized as follows. In the section II, we detail the architecture of the skyrmion reservoir and corresponding magnetization dynamics, in section III we describe the process to set up and train and test the reservoir and in section IV we discuss the results and summarize our findings in the conclusion (section V).

## II. METHODOLOGY

### A. PROPOSED PHYSICAL RESERVOIR

Fig. 1a shows a conventional reservoir computing system which consists of an input layer, a reservoir block having recurrent connections among the constituent nodes and an output layer. The solid line arrows show the connections which are fixed and the dashed arrows are the connections which need to be trained. We propose to replace the reservoir block with three patterned and decoupled skyrmions whose magnetization dynamics we simulate. Each of the individual skyrmions is hosted in the ferromagnetic thin films with perpendicular magnetic anisotropy (PMA) as shown in Fig. 1b. A ferromagnetic reference layer, a tunnel barrier (MgO) and a synthetic antiferromagnetic (SAF) layer are patterned on top of the ferromagnetic free layer (that hosts the skyrmion) to create the MTJ as shown in Fig. 1c. This facilitates the read and write operation. The temporal inputs are linearly mapped into a voltage pulse and applied across the MTJs to modulate the PMA using the VCMA effect [52-54]. All the patterned skyrmions are subjected to the same set of inputs. When the PMA is modulated within a certain range, the skyrmions generate oscillatory response (skyrmion breathing) as shown in Fig. 2. The responses of the skyrmions are read with MTJs and processed with linear regression to compute the predicted values of the temporal time series.

For a typical reservoir consists of N number of nodes as seen in Fig. 1a, the time discretized states of the nodes, $r_i^n$, can be represented as follows:

$$r_i^{n+1} = f(\sum_{j=0}^{N-1} p_{ij} r_j^n + q_i u^n) \quad (1)$$

Here, the $p_{ij}$, $q_i$ are time-independent coefficients that are drawn from a random distribution having a mean of 0 and the standard deviations are adjusted for optimal performances. Also, $f$ is the activation function which can be linear or non-linear and $u^n$ is the input. Here, the "fading memory" or the short-term memory (an essential property of the reservoir) is achieved by using large number of nodes and their recurrent connections. In comparison, the skyrmion systems have inherent memory effect in their responses, thus instead of using several nodes only one skyrmion device can work as a reservoir. However, to increase the dimensionality, the states of this reservoir can be read at regular interval for a particular input, which acts as the virtual nodes of the reservoir. The concept is originally developed in reservoir with delayed feedback where a nonlinear node subjected to an input and delayed feedback acts as a chain of virtual nodes and provide performance similar to a typical reservoir [45]. Later on, it has been shown the virtual nodes derived from reservoir responses subjected to only the input signal can provide optimal performance [17,18]. In such a scenario, if the virtual node interval (as shown by $\theta$ in Fig. 1d) is lower than the characteristic time (relaxation time) of the reservoir, the node states are not only influenced by their own previous states, but also the neighboring node states and the input excitation. This allows for non-linear coupling among the nodes.

Thus, the interconnection matrix in Eq. 1 is simplified in the skyrmion reservoir case where the virtual nodes are assumed to be connected in ring topology (as seen in Fig. 1d). The resulting node states of the reservoir can be expressed as:

$$r_0^{n+1} = f(r_0^n + r_{N-1}^{n-1} + u^n)$$
$$r_i^{n+1} = f(r_i^n + r_{i-1}^n + u^n) \quad (2)$$

Here, we have used linear activation function, $f(w)=w$, thus the read-out reservoir states are used for training without any post-processing or non-linear transformation.

The node states of the reservoir act as features, which are used to generate the output. Since we are using only linear activation, we also include several previous responses of the reservoir for generating the output. Due to the short term-memory effect inherent to the skyrmion dynamics, these previous states provide additional non-linear effects. The output of the reservoir can be expressed as follows:

$$y^n = \sum_{j=n-d}^{n} \sum_{i=0}^{N-1} w_{ij} r_i^j \quad (3)$$

Where, d represents the number of time-steps for which the previous reservoir responses are included. The optimal weights, $w_{ij}$ can be obtained by optimizing a cost function. We use mean squared error as our cost functions:

$$c = <(y^n - t^n)^2> \quad (4)$$

Where $t^n$ represents the teacher or target output of the system at $n^{th}$ time step. The optimization can be performed off-line using linear regression with regularization (ridge regression) or on-line using gradient descent optimizer.

During the training or weight optimization stage, the teacher input, $I^n = u^n$ is applied to the reservoir to predict the next time step value $y^n = u^{n+1}$ as seen from Fig. 1d. Once the optimized weights are obtained, the testing phase begins, where the inputs are disconnected and the output of the reservoir is connected directly to the input, $I^n = y^{n-1}$ as can be seen in Fig. 1e.

### B. MAGNETIZATION DYNAMICS

The magnetization dynamics of the skyrmions induced by PMA modulation in the thin film is simulated by solving the Landau-Lifshitz-Gilbert (LLG) equation [55,56] using the MUMAX3 simulation package [57]:

$$(1 + \alpha^2)\frac{d\vec{m}}{dt} = -\gamma \vec{m} \times \vec{H}_{eff} - \alpha\gamma \left(\vec{m} \times (\vec{m} \times \vec{H}_{eff})\right) \quad (5)$$

Where $\gamma$ and $\alpha$ represent the gyromagnetic ratio and the Gilbert damping coefficient respectively. $\vec{m}$ stands for the normalized magnetization vector, which is found by normalizing the magnetization vector ($\vec{M}$) with respect to saturation magnetization ($M_s$). The thin films are discretized into cells with dimensions of 2 nm × 2 nm × 1 nm, which are much shorter than the exchange length ($\sqrt{\frac{2A_{ex}}{\mu_0 M_s^2}}$). In equation (5), the effective field, $\vec{H}_{eff}$ accounts for the contributions from PMA, demagnetization, exchange interaction arises from Heisenberg interaction [58] and Dzyaloshinskii–Moriya interaction (DMI) [59]. $\vec{H}_{eff}$ can be expressed as follows [57]:

$$\vec{H}_{eff} = \vec{H}_{anis} + \vec{H}_{demag} + \vec{H}_{exch,Heisn} + \vec{H}_{exch,DMI} \qquad (6)$$

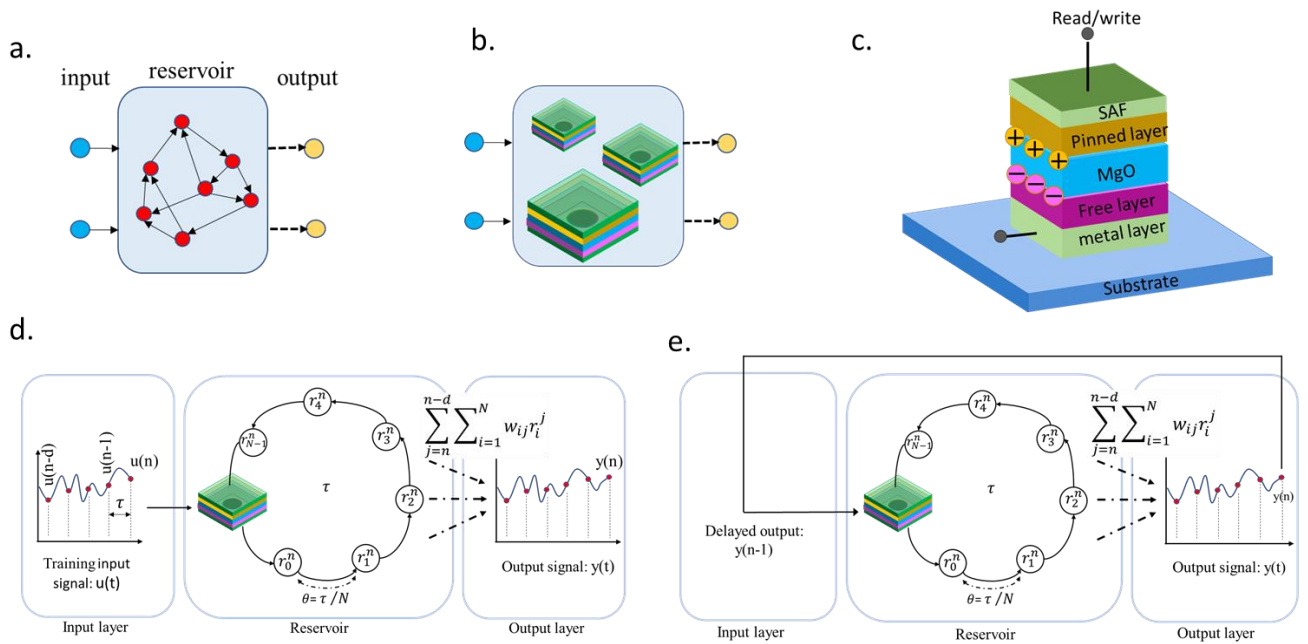

FIGURE 1. a. A conventional reservoir computing system with input layer, reservoir block with recurrent connections among nodes and the output layer. b. The reservoir block is replaced by a set of patterned skyrmion devices where each of the ferromagnetic films with PMA host a single skyrmion. c. Stacks of a skyrmion device with metallic electrode and MTJ. d. Training of a skyrmion reservoir for autonomous prediction task. The temporal input data is mapped into voltage values which are applied to each of the skyrmion devices and the responses are collected. The responses are read at regular interval and the read-out values act as the virtual node as represented by $r_i^j$. The state of the nodes (or reservoir responses) are used to predict the next time step value of the input time series. The weights are trained by computing the error of the predicted and target values and accomplished with simple pseudoinverse operation. e. During testing, the predicted output value is directly fed as input to the reservoir in order to perform multi-step autonomous prediction.

$\vec{H}_{exch,DMI}$ is the effective field due to Dzyaloshinskii-Moriya interaction which is defined as [59]:

$$\vec{H}_{exch,DMI} = \frac{2D}{\mu_0 M_s}\left(\frac{\partial m_z}{\partial x}, \frac{\partial m_z}{\partial y}, -\frac{\partial m_x}{\partial x} - \frac{\partial m_y}{\partial y}\right) \qquad (7)$$

Where D is the DMI constant and $m_x$, $m_y$, and $m_z$ are the components of unit magnetization vector $\vec{m}$ along x, y, and z direction, respectively.

$\vec{H}_{anis}$ is the effective field due to the perpendicular anisotropy and expressed by the following equation [57]:

$$\vec{H}_{anis} = \frac{2K_u}{\mu_0 M_s}(\vec{u}.\vec{m})\vec{u} \qquad (8)$$

Where $K_u$ is the first order uniaxial anisotropy constant with $\vec{u}$ is a unit vector along the anisotropy direction. Applying the voltages across the MTJ changes the PMA, which is obtained by changing the $K_u$ values. We note that, PMA (or $K_u$ coefficient) can be increased (decreased) by applying a negative (positive) voltage [34,60]. Applying a positive (negative) voltage decreases (increases) the energy barrier exists between the perpendicular and in-plane magnetized states of the free layer. When the barrier is perturbed, the skyrmion undergoes oscillatory response which eventually settles down after some time (a few hundred nanosecond) if the perturbation is kept small (large enough perturbation can switch states and not desired). The simulations have been carried out without the thermal noise (T=0 K), however, from our previous study it has been shown that the short-term memory property of the skyrmion reservoir does not degrade much in the presence of room temperature thermal noise [31]. The simulation parameters are listed in Table I.

## C. DATASET
We evaluated the long-term prediction performance of our proposed reservoir on two different time series forecasting datasets.

TABLE I
SIMULATION PARAMETER

| Parameters | Values |
| --- | --- |
| DMI constant (D) | $0.00065\ Jm^{-2}$ |
| Gilbert damping ($\alpha$) | 0.01 |
| Saturation magnetization ($M_s$) | $1.3 \times 10^6\ Am^{-1}$ |
| Exchange constant ($A_{ex}$) | $15 \times 10^{-11} Jm^{-1}$ |
| Perpendicular Magnetic Anisotropy ($K_u$) | $10.925 \times 10^5\ Jm^{-3}$ |

### 1) MACKEY-GLASS TIME SERIES
The MG time series can be expressed as non-linear time-delay differential equation as follows:

$$\frac{dx}{dt} = \sigma\frac{x(t-\tau)}{1+\left(x(t-\tau)\right)^n} - \beta x(t) \qquad (9)$$

where $x(t)$ is the MG time series value at time $t$, and $\sigma$=0.2, $\beta$=0.1, n=10, $\tau$=17 and $x$ (0) =1. The equation is solved using Runge-Kutta method with integration time step, $dt$=0.1. The time series is downsampled to 10 and normalized between [-1,1]. Despite the deterministic form, forecasting this chaotic series is challenging, thus the system is used for benchmark forecasting tasks in literature [42-44]. 431-time steps data are considered where the first 30-time steps data (1-30) are not trained, next 370-time steps data (31-400) are trained and the next 30-time steps data (402-431) are predicted with autonomous prediction.

### 2) INDIVIDUAL HOUSEHOLD POWER CONSUMPTION
The other dataset we worked on is a benchmark dataset of electricity consumption for a single residential customer, named "Individual household electric power consumption" [61]. The data set contained power consumption measurements gathered between December 2006 and November 2010 with a one-minute resolution. The dataset contained aggregate active power load for the whole house and three sub-metering for three sections of the house. In this paper, only the aggregate active load values for the whole house are used. The dataset contained 2075259 measurements. The hourly resolution data were obtained by averaging the one-minute resolution data. Multistep autonomous prediction is especially challenging for these types of datasets due to the stochasticity in the data that arises from erratic human behavior or seasonal change. 284 hours of data are considered where, the first 20 hours data (1-20) are not trained, the next 240 hours of data (21-260) are used for training and the final 23 hours of data (262-284) are predicted with autonomous prediction.

## III. RESERVOIR SETUP

### A. SKYRMION RESERVOIR

Three patterned ferromagnetic thin films with square geometry having the side lengths of 1000 nm, 800 nm and 700 nm are considered as the reservoir for MG time series prediction tasks as shown in Fig. 2. 15 nm thick slices are etched from all sides in the middle region, which leaves a 500 nm square block hosting the skyrmions. The etched block is used to prevent the propagation of spin waves, which is shown to negatively impact the short-term memory capacity of the skyrmion reservoir [31]. Moreover, the etched region will provide a boundary so that the skyrmion cannot be annihilated easily. In addition, different length scales of the periphery will provide different strength of dipole coupling to the skyrmions and create variability in the reservoir response thus enhancing the robustness of the reservoir. We note, input multiplexing is used in previous studies to incorporate diverse response of a single reservoir node. However, we do not use input multiplexing and the variation of responses is incorporated by using different geometry nodes. The input time series values are transformed into voltage pulses with linear conversion between input magnitude and voltage applied. This voltage pulse translates to change in the perpendicular magnetic anisotropy using the VCMA coefficient, $\varepsilon = \frac{\Delta K_{si}}{\Delta V/t_{MgO}}$ (described later in section IV) in the skyrmion devices. For modeling the response, we use voltage pulses which are applied sequentially with a 2 ns duration. After applying each input pulse, the system is relaxed for 16 ns. We note that instead of applying the pulse for 18 ns we opt for a shorter write pulse which not only saves energy but also triggers rich dynamics that occurs during the relaxation phase of the skyrmion device. Moreover, the relaxation offers flexibility in terms of post processing time required for multi-step autonomous prediction (current prediction is provided as input for the next prediction). Fig. 2 shows the magnetization responses of the different reservoirs during the training phase of the MG time series at time step 231-235. The reservoir responses for a single period (18 ns) is read at 3 ns interval (6 times). These 6 values act as the virtual nodes of the reservoir (see in Fig. 2, the red diamond marks). Nodes more than 6 can be selected; however, this does not improve the performance as 6 nodes can adequately capture the amount of information in one period. Next, instead of using only the states in the current period (as been done in our previous study for one step ahead prediction [27,31]), we also include reservoir states from the previous 30 periods of data (total 31 periods, 31*6=186 states for one skyrmion device) for the autonomous long-term prediction of the MG series. The short-term memory capacity of the patterned skyrmion is shown to be ~ 4 bits [31] (up to 4 periods). Thus, the reservoir is expected to remember inputs from the previous 4 periods and non-linearly transform its states based on the memorized inputs. However, when tasked with autonomous prediction using only the current states, the prediction quickly diverges (large prediction errors after third time-step). Thus, for multi-step prediction, previous responses are included to enable the RC to utilize important contexts from the past few observations. Once the reservoir states are obtained from all three skyrmion devices, ridge regression (Tikhonov regularization) is performed for training and computing the optimal weights. The mean squared error is considered as the cost function (shown in Eq. 4) and the activation functions for the reservoir nodes are considered to be linear. With these assumptions, the optimal weights can be found using the following:

$$w_{ij}^{opt} = (AA^T + \lambda I)^{-1} A^T B \qquad (10)$$

where, A is the reservoir response vector including the present and past observation for all the training inputs, $\lambda$ is the regularization coefficient, B is the vector of labels containing the target output. For MG series forecasting we choose the regularization coefficient to be, $\lambda = 10^{-8}$.

For household power prediction tasks, slightly modified geometries are used. The ferromagnetic thin films of 1050 nm, 850 nm and 750 nm are used. The additional side lengths of the ferromagnetic regions provide stability to the skyrmions to the stochastic changes presented in the household load data. The etched region of 15 nm and 500 nm middle regions for hosting the skyrmions remain the same. The voltage pulse duration of 2 ns and relaxation period of 16 ns are kept the same. Here, the reservoir states for a total of 21 periods (current period, and previous 20 periods) are used for forecasting tasks. The optimal weights are computed using ridge regression where the regularization parameter is chosen to be, $\lambda = 10^{-1}$. Due to nonvolatile and stochastic nature of the household load data, it is difficult to train the reservoir as accurately as possible without overfitting, thus a large regularization coefficient is required.

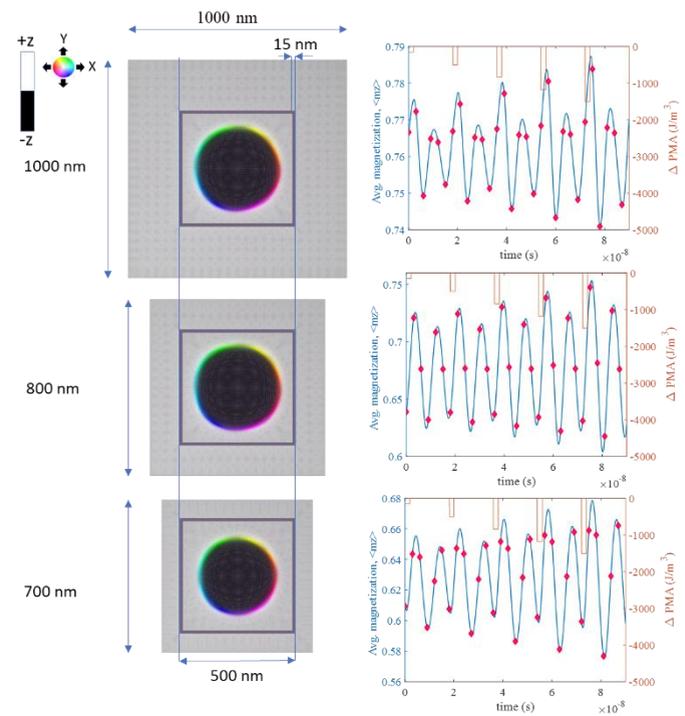

**FIGURE 2.** a. Three ferromagnetic thin films each hosting a magnetic skyrmion worked as the RC. The responses of the respective skyrmion devices are shown side by side when the thin films are perturbed by the inputs of MG time series from time-step 231 to 235 (inputs are mapped into voltage pulse amplitudes). The PMA modulation by the input voltage pulses are shown in orange color. The virtual nodes are marked in red diamond.

### B. TRAINING AND TESTING THE RESERVOIR

At first, for each of the input, the reservoir states are read at 3 ns interval up to 18 ns. The state vector can be expressed as $R_n = \{r_0^n, r_1^n, \ldots, r_5^n\}$, where the superscript $n$ in $r$ represents the n[th] input of the temporal series and the subscript represents the virtual node number. For MG series prediction task, all 400 training inputs are applied sequentially to all of the skyrmion devices and the corresponding $R_n$ are collected. The label, $t^n$ for the prediction task is the next time-step MG function value, $t^n = u^{n+1}$, where $u^n$ is the MG function value at n[th] time step. The training could be performed using the reservoir response vector, A= $[(R_1, \ldots, R_d, R_{d+1})', \ldots, (R_{n-d-1}, \ldots, R_{n-2}, R_{n-1})', (R_{n-d}, \ldots, R_{n-1}, R_n)']$ and the corresponding target output vector, B= $[t^1, \ldots, t^{n-1}, t^n]$ and using the ridge regression equation in Eq. 6. The "'" symbol denotes the transpose operation. Once the optimal weights are obtained after training the reservoir is ready for testing and performing autonomous prediction.

In an actual hardware implementation, the weight optimization (such as pseudoinverse) operation can take time. By the time the optimal weights are computed, the reservoir can be sufficiently relaxed and loses its memory. Thus, before starting the testing phase, the reservoir needs to warm up [17,42]. The same temporal series used in training can be used for warm-up or initializing the reservoir. This adds computational overhead to the model. However, during such initialization step the reservoir is only excited with actual training data and reading the states of the reservoir is not required, which saves read energy cost. The overhead of the reservoir initialization can be avoided by allocating some of the training data for initialization. Depending on the postprocessing optimization time, instead of using all the training data for weight optimization, some of the training data can be saved for reservoir initialization. Alternatively, the optimal weights can be obtained using simple gradient descent method, which can be performed at the same time the training data are supplied to the physical reservoir. This is demonstrated by authors in their opto-electronic reservoir implementation in FPGA [62]. Furthermore, the time complexity of matrix-vector multiplication (the most computationally expensive load of stochastic gradient decent optimization) could be further reduced to single time step using non-volatile computational memory device arranged in a crossbar. In the MG series prediction task, during the warm-up, the inputs are applied sequentially up to 401[th] input. Once the reservoir responses are collected, we predict the 402[th] time step value of the series and then this predicted output is fed directly to the reservoir as input as shown in Fig. 1e. We repeatedly performed these steps to autonomously predict the time series up to 431[th] time step. For autonomously predicting household power consumption, the reservoir is trained and initialized by providing input up to 261[th] hour data. Then, the reservoir responses are collected and using the optimal weights, the 262[th] hour data is predicted. The predicted output is then directly fed as the input. Autonomous prediction up to 284[th] hour is performed following these steps.

### C. SEQUENCE-TO-SEQUENCE LSTM ARCHITECTURE

The performance of the reservoir is also compared with state-of-the-art sequence-to-sequence (S2S) LSTM. S2S is an architecture that was proposed to map sequences of different lengths [63]. The architecture consists of two LSTM networks: an encoder and a decoder. The encoder's job is to transform input sequences of variable length into fixed-length vectors, which will then be used as the input state of the decoder. If we assume, the decoder produces an output sequence with length $l$. Then, for this instance, the decoder output is the energy load projection for the following $l$ time steps. This architecture's key benefit is that it accepts inputs of any length. Thus, electricity load for an arbitrary number of future time steps can be predicted using any number of available load measurements of previous time steps as inputs. Given, $M$ historical electricity load measurements available, which can be expressed as:

$$y = \{y[0], y[1], \ldots, y[M-1]\} \quad (11)$$

where $y[t]$ is the actual load measurement for time step t. The load for the following $T - M$ time steps should be predicted. The predicted load values can be expressed as;

$$\hat{y} = \{\hat{y}[M], \hat{y}[M+1], \ldots \hat{y}[T]\} \quad (12)$$

For training, the encoder network is pre-trained to minimize the following error:

$$LE = \sum_{i=1}^{M}(y[i] - \hat{y}[i])^2 \quad (13)$$

Then the encoder is plugged into the decoder network and both of the networks are trained to reduce the objective function:

$$LD = \sum_{i=M+1}^{T}(y[i] - \hat{y}[i])^2 \quad (14)$$

The error of network is minimized using the backpropagation algorithm. Back-propagation signals are allowed to flow from the decoder to the encoder. Therefore, weights for both the encoder and decoder are updated in order to minimize the objective function expressed in Eq. 14. Both decoder and encoder are updated because the pre-training of the encoder alone is insufficient to achieve good performance. In this paper, we tested multiple layers with different numbers of neuron units per layer and we found that the training dataset using a 2-layer network with 50 units in each layer gave the best performance. Increasing the capacity of the network did not improve performance on the testing data.

## IV. RESULTS AND DISCUSSIONS

### A. AUTONOMOUS PREDICTION WITH RESERVOIR

Long-term prediction results of the proposed reservoir for MG time series prediction is shown in Fig. 3a. After training the reservoir up to 400 time-step data, the input is disconnected and the predicted output is connected to the reservoir input. The reservoir is able to predict the next 30 time-step output with very good accuracy and with a root mean squared error (RMSE) of 0.0015. As the errors after 30-time steps prediction is extremely small, further prediction is possible. However, we restrict our effort due to the simulation complexity and limited hardware resources. Fig. 3c and Fig. 3d show the phase plot of the training and testing data of the chaotic MG attractor.

The overlapping plots in Fig. 3c and 3d show that the reservoir was able to accurately predict both of the training and test data. When the same task is given to the LSTM as shown in Fig. 3b, it performs well and the resulting RMSE was 0.0000013. This performance improvement can be due to the dependencies that arise among the LSTM cell states in the 2-layer deep architectures because of the use of forget gates (control how many previous states to remember). In addition, the non-linear activation functions are used for the input, output and forget gates which provide the non-linear transformation effect. Despite using much simpler architectures and linear activations, the reservoir is able to predict the chaotic trend with competitive accuracy. In our reservoir, during testing, the output is directly fed as the input, and this connection is not scaled or optimized (as has been done in [42]). Moreover, we did not use any non-linear transformation of the reservoir states, rather use the states as it is, and included several previous states. Thus, the success of the reservoir for long term prediction can be attributed to the use of previous reservoir states during the training and testing which provides the necessary non-linearity. Interestingly, here the non-linearity arises from the physical reservoir's dynamic responses rather than using any external non-linear transformation (trigonometric or polynomial activation). Furthermore, we did not use any temporal mask to encode an input for generating diversified features, instead we used different geometry skyrmion devices for variability. The only parameter that is scanned and optimized during training is the number of previous reservoir states. For MG prediction, we used reservoir states for 30 previous inputs, which is shown to provide adequate non-linearity. For long term prediction with output feedback, it is extremely important to predict the immediate future steps as accurate as possible, as any error in the near future can accumulate fast and make the prediction divergent. Previously, reservoir with delayed input is shown to improve the one-step ahead prediction of the reservoir [45], which demonstrates the importance of the non-linearity effect coming from the delayed input to the prediction performance. In our proposed reservoir, accuracy of the long-term prediction is maintained due to the inclusion of previous states (similar effect as of the delayed input).

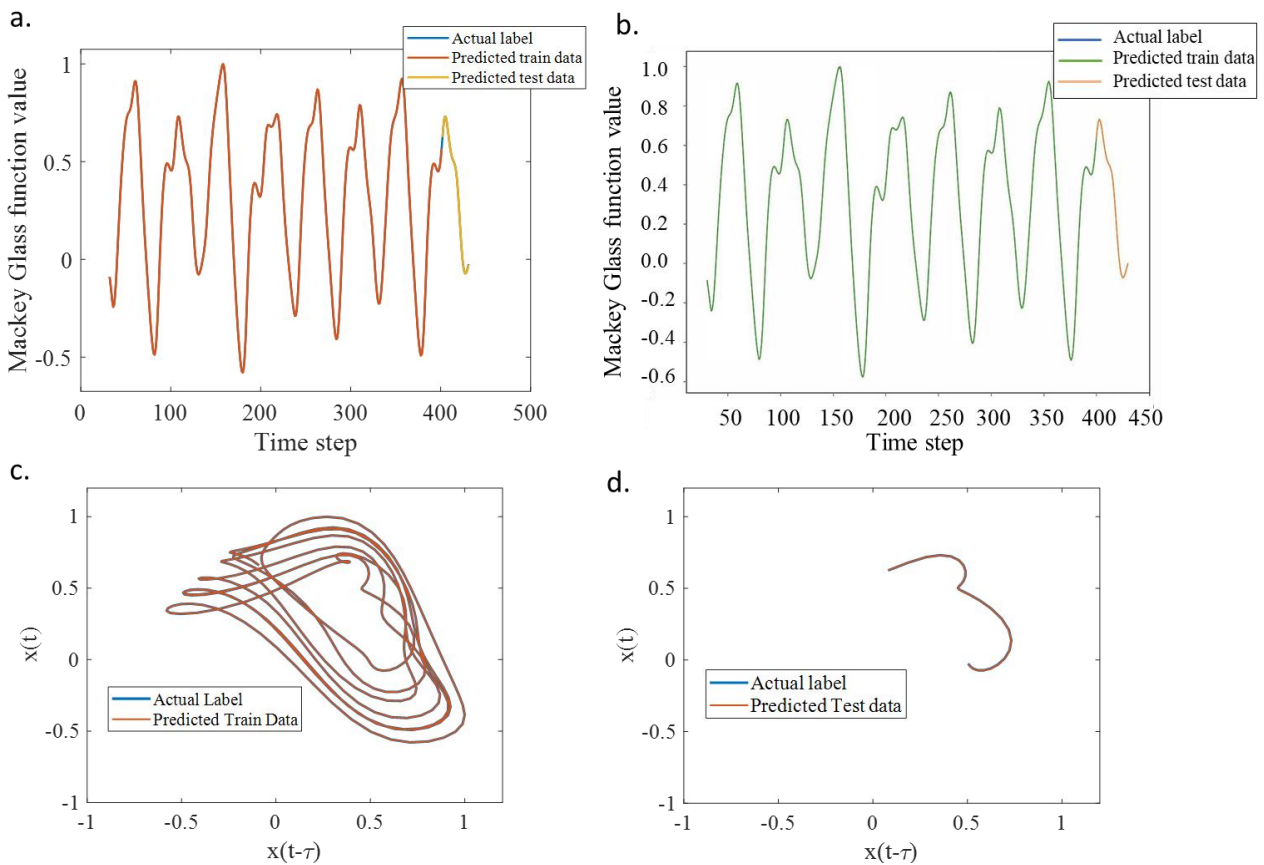

**FIGURE 3.** a. Long term autonomous prediction of chaotic MG time series with skyrmion reservoir. The dataset is trained with 31-400 time-step data. The reservoir is tasked to predict the next 30-time step data from 402-431. The overlapping of the predicted test data with actual label suggests accurate prediction b. prediction trend for MG time series with 2-layer deep sequence to sequence LSTM architecture. The LSTM is able to accurately predict the trend. Although, RMSE magnitude of the LSTM is lower than the reservoir, however the prediction errors for both of the predictions remain extremely small. c. Phase diagram of the chaotic MG attractor during the training with reservoir. The predicted training data is overlapped with actual label implying the efficacy of the ridge regression training. d. Phase diagram of the reservoir for autonomous prediction. The superimposed plots suggest good prediction accuracy of the reservoir on test data.

After the successful performance of the proposed reservoir for the long-term chaotic time series prediction task, we focus on forecasting the long-term individual household power consumption. The task is challenging due to the non-volatile and univariate nature of the data (only the power value is readily available, other parameters are unknown) and especially when the dataset is small. However, we find that the proposed skyrmion reservoir can achieve good accuracy when we use several previous reservoir responses. The total number of previous states that is included in the training are optimized and reservoir states for 20 previous inputs are used. The autonomous long-term prediction results are presented in Fig.

4a and 4b for the proposed reservoir and 2-layer deep S2S LSTM architecture respectively. From Fig. 4a, it is clear that the reservoir is able to predict the household power demand with good accuracy. The RMSE after 23 hours of prediction is calculated to be 0.0885. The prediction accuracy of the reservoir is good in the first several hours of the prediction (see the hourly RMSE plot in Fig. 5 labeled as reservoir: 262-284). In contrast, the accuracy of the LSTM is poor at the beginning of the prediction, however, regains accuracy in the next few predictions and the overall RMSE is calculated to be 0.0831. The accuracy degradation of the LSTM for the first few predictions could be due to lack of training data, as LSTM typically requires large numbers of observations to find the underlying dependencies in the data.

Further, the reservoir is tasked to predict the next 286-308 hours of data where the load consumption trend is significantly stochastic. The hourly RMSE for both the reservoir and LSTM prediction for two different intervals: 262 to 284 hours and 286 to 308 hours are presented in Fig. 5. The training and testing trend of the reservoir for 286 to 308 hours interval is shown in the inset of Fig. 5. From the RMSE plot we can see that, in 286 to 308 hours interval, the initial prediction of the reservoir is beyond the magnitude of the regularization ($\lambda=0.1$), however, the reservoir is able to minimize the error in the next few predictions and follow the load trend. The large prediction error of the reservoir for the 286-308 hours of load compared to the previous set of prediction (262-284 hours) arises due to the significantly stochastic load behavior which is difficult to track with the reservoir. The limitation of the reservoir to track unstable stochastic trend is also observed in previous study [41].

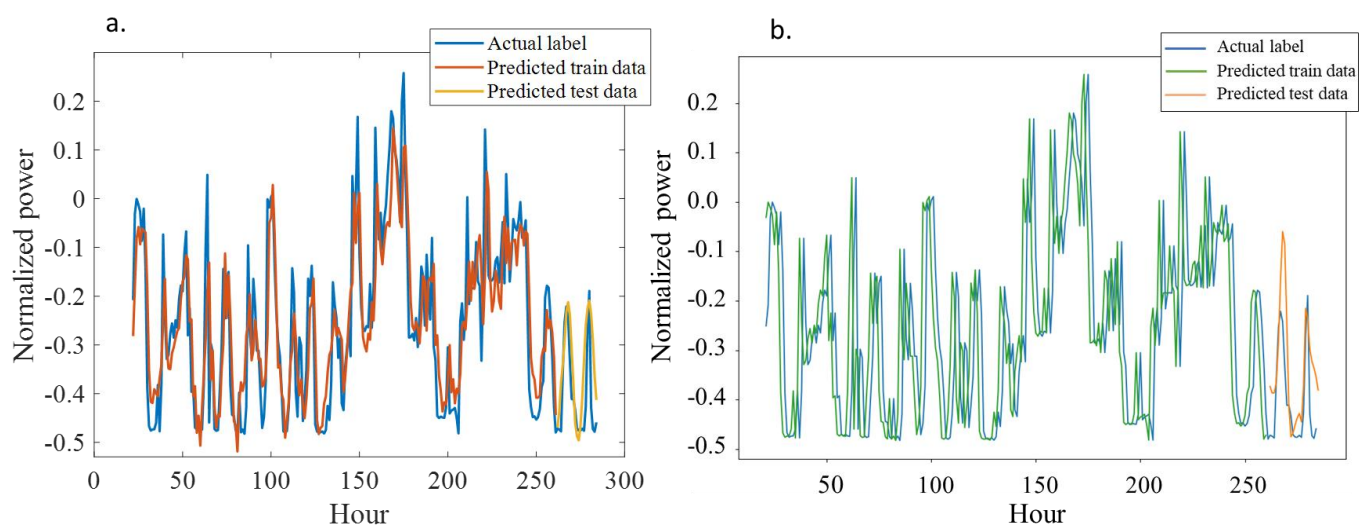

FIGURE 4. a. Long-term autonomous forecasting of individual household active power load with proposed reservoir. The reservoir is trained with 21-261 hours of data and tasked with predicting the next 23 hours of data. The predicted trend closely follows the actual load level suggesting good prediction accuracy with the skyrmion reservoir. b. The same task is performed with 2-layer sequence to sequence LSTM architecture. Although, the LSTM is able to capture the trend, the prediction accuracy is less than the proposed reservoir for the first several hours of prediction.

For both of the prediction intervals, the reservoir prediction accuracy starts to diverge after several hours. As mentioned earlier, the long-term prediction with feedback depends on the accurate prediction of the immediate future. However, due to nonvolatile and stochastic trend of the household load data, the prediction accuracy degrades quickly, nonetheless good accuracy is found up to 20 hours. Compared to the reservoir, the LSTM prediction accuracy is poor for such stochastic load trends. The steady and higher prediction accuracy of the reservoir compared to the LSTM further proves the efficacy of the RC for the smaller dataset.

Finally, the RMSE of our proposed RC is compared with recent transformer [64] and generative recurrent unit (GRU) based optimization algorithms [65]. The RMSE of the RC for household power prediction task is computed to be ~ 0.6 kW and the LSTM sequence to sequence model RMSE is 0.66 kW [51]. The best performance with regards to RMSE for the selective update and adaptive power optimization based GRU method is shown to be ~ 0.15 kW [65]. On the other hand, the sparse transformer-based model with adversarial network learning-based approach [64] has RMSE ~ 0.30 kW (~ 50% accuracy improvement in terms of RMSE in GRU based approach [65]). Although GRU and transformer-based approach shows higher prediction performance, they incur huge computational burden and associated energy cost for training. The computational complexity of the self-attention (most data intensive task in transformer) is quadratic as such the complexity for each position of a transformer can be expressed as O (d.$N^2$) where $d$ is the dimension of each position of a sequence length of $N$. Additional complexity comes from feedforward linear operation, positional encoding, layer normalization. In GRU network based adaptive optimization approach, the hidden state information is compared at each time step which incurs an additional complexity of O ($d$) on top of the GRU unit's computational complexity of O ($N.k.d$), where d is the dimension of hidden state vector and $k$ is the number of GRU unit. Furthermore, to incorporate temporal dependency in gradient flow, the author in [65] proposes an additional hyperparameter called memory factor which requires additional multiplication and addition operation compared to conventional optimizer. Furthermore, the optimal value for the hyperparameter needs to be scanned using grid search, which is a computationally extensive process. In comparison, the reservoir states are evolved by the internal dynamics of spintronic device. Thus, only a single feedforward layer needs to be trained to predict the future trajectories of the time-series. For a P × 1 layer feedforward network (P is the number of reservoir internal states that are read, and the RC output is a single prediction), the computational complexity can be expressed as O (P) which is significantly smaller than the transformer, GRU and LSTM based networks. Thus, physical RC based approach could be a viable solution for edge intelligence especially in the scenario where we can trade-off accuracy in favor of limited resources.

### B. ENERGY DISSIPATION

There are two main contributions to the energy consumed for the reservoir computing discussed here. First, energy is needed to modulate the PMA of the ferromagnetic layers. Maximum change in PMA coefficients is, $\Delta PMA = 7.5 \times 10^3 \, J/m^3$. Thus, the maximum change in the surface anisotropy coefficients is estimated to be, $\Delta K_{si} = \Delta PMA . t_{CoFeB} = 7.5 \times 10^3 . 1 \times 10^{-9} = 7.5 \times 10^{-6}$. Assuming the VCMA coefficient of the MTJ to be, $\varepsilon = \frac{\Delta K_{si}}{\Delta V/t_{MgO}}$=31 fJ/V-m [66], the thickness of the tunneling barrier MgO to be, $t_{MgO}$=1 nm, the magnitude of voltage to perform the maximum PMA modulation can be calculated to be, $\Delta V$=0.24 V. Assuming the relative permittivity of MgO to be 7, the total capacitance can be calculated to be, C=$\frac{\epsilon_0 \epsilon_r (L*L)}{t_{MgO}}$ ~62 fF. Here, we assume L=1050 nm (so our estimate is conservative) for the length of the side of the square region of ferromagnetic layer. Thus, the total write energy to charge the capacitive tunneling region is estimated to be $\frac{1}{2}CV^2$~ 2 fJ. Second, the reservoir responses that are read after certain interval (3 ns) known as virtual nodes also consumes energy. The read energy can be estimated to be ~1.24 fJ with a read delay of ~ 0.3 ns [67] (which is well within 3 ns interval). In a period of 18 ns, the read operation is performed 6 times. Thus, the total energy for the write (PMA modulation) and read energy is calculated to be ~ 9.44 fJ. For the household prediction task, a total of 21 hours of data (which translates to 21 discrete data points for the reservoir computing) are used to predict the next-hour household power. Thus, the total energy dissipation for one skyrmion reservoir is = 21*9.44~ 198 fJ. During the reservoir initialization stage, the PMA is modulated. However, the states are not read. The total energy during the reservoir initialization stage is estimated to be ~ 440 fJ. Including the reservoir initialization energy, the total energy consumption for 3 skyrmion reservoir to predict one future value of individual household energy consumption is estimated to be ~ 651 fJ (assuming worst case scenario). For LSTM implementation, the GPU energy is calculated to be ~0.68 J per prediction. With reservoir implementation, the output layer is a feedforward layer that is implemented in a GPU as well for fair comparison. The GPU energy consumption for the reservoir feedforward layer is calculated to be 0.043 J per prediction. Thus, the reservoir is able to predict one instance with 16× lower energy compared to the LSTM. We note that further reduction in energy consumption with reservoir can be achieved by implementing feedforward output layer with crossbar array of non-volatile memory using in-memory computing [68,69]. Thus, we could get an overall (at the system level which is the key metric of interest) 10× to 100× reduction in the computing energy needed to predict building energy. At the individual RC PMA device level, the energy savings would be enormous (pico-Joules instead of milli-Joules) but we do not use this metric as it is not a fair comparison if overall architecture and systems are not considered.

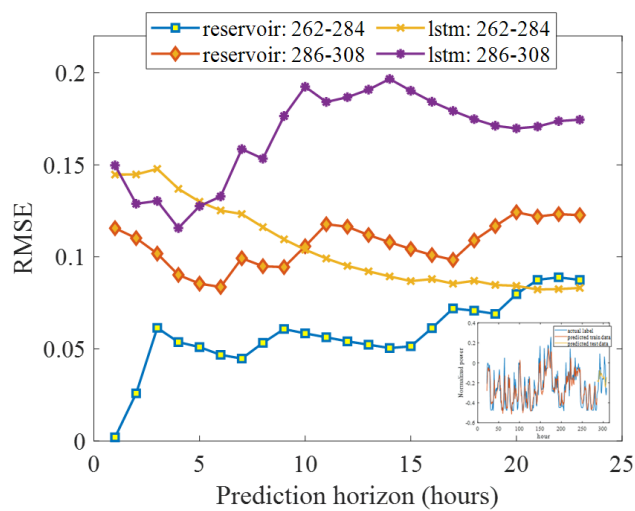

**FIGURE 5.** a. Hourly RMSE of the prediction accuracy for individual household load forecasting task for both of the proposed reservoir and LSTM. RMSE plots for two different long-term autonomous predictions, 262-284 hours and 286-308 hours are shown. The inset shows the prediction trend of the reservoir for prediction from 286 hour to 308 hours. The RMSE plots indicate higher prediction accuracy of the proposed reservoir compared to LSTM, even for much stochastic trend such as in 286-308 hours of data.

## V. CONCLUSION

We have shown long-term autonomous prediction with a skyrmion reservoir. The reservoir is tasked with predicting the chaotic MG time series and real-world individual household load forecasting and is able to predict long-term trends with competitive accuracy. The proposed reservoir is set up using three patterned skyrmions having slightly different geometries. All the skyrmions are provided with the same temporal input series and the resulting skyrmions' oscillation is read at regular intervals and processed with simple linear regression. After training, the output is fed as the reservoir input to perform autonomous long-term prediction. The prediction performance greatly improves due to inclusion of previous states in addition to the current states of the reservoir as these previous states provide the non-linear effect necessary for accurate prediction. Furthermore, the physical reservoir does not consider masking the input and only the output weights and the number of previous states included during training are optimized. Energy consumption estimation shows that skyrmion reservoir can perform autonomous prediction with an energy consumption of 0.043 J/per prediction which is at least 16× less than the LSTM based approach. In addition, we show that with our proposed physical RC one can achieve competitive accuracy with much smaller dataset. Furthermore, with VCMA control, the skyrmion reservoir can be operated with ultra-low power as the anisotropy modulation is performed with voltage as opposed to energy hungry current control. Since in RC only the last layer is trained, thus our skyrmion reservoir provides a pathway to implement extremely energy efficient long-term prediction of real-world problem with high accuracy, which is specifically attractive in hardware and memory constraint edge computing platforms, where energy is at a premium.

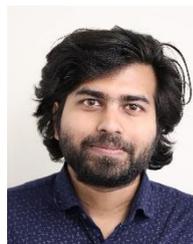

**WALID AL MISBA** received the B.Sc. degree in electrical and electronic engineering from the Bangladesh University of Engineering and Technology, Dhaka, Bangladesh in 2013. He received the MS in electrical engineering from Tuskegee University, Alabama, US. He is currently pursuing his Ph.D. degree in Mechanical and Nuclear Engineering with Virginia Commonwealth University, Richmond, VA, USA

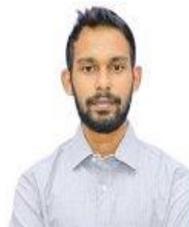

**HARINDRA S. MAVIKUMBURE** received his B.Sc. degree in computer science from the University of Peradeniya, Sri Lanka, in 2018. He is currently working as a research assistant while reading for his doctoral degree in computer science at Virginia Commonwealth University, Richmond. His research interests include Anomaly Detection, Machine Learning, Deep Learning, and Unsupervised Learning.

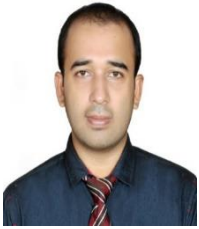

**MD MAHADI RAJIB** received the B.Sc. degree in mechanical engineering from the Bangladesh University of Engineering and Technology (BUET), Dhaka, Bangladesh in 2016. He received the MS in mechanical and nuclear engineering from Virginia Commonwealth University (VCU), Richmond, VA, US and currently pursuing his Ph.D. degree in the same departmen

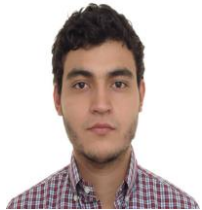

**DANIEL L MARINO** (S'14) received his PhD in engineering from Virginia Commonwealth University, USA, in 2021. He received his B.Eng. in automation engineering from La Salle University, Colombia, in 2015. He is currently a research assistant and a doctoral candidate at Virginia Commonwealth University, with over six years of research and development experience, collaborating with US DOE National Labs, universities, and industry partners. He has authored over 27 articles in peer reviewed journals and conferences. He received the IEEE IES student paper travel award in 2016 and 2019, the VCU CS Outstanding Paper Award in 2020, the VCU CS outstanding early-career student researcher award in 2017, and the honor scholarship granted by La Salle University from 2010 to 2013. His research interests include stochastic modeling, deep learning, and explainable AI with applications in cyber-physical systems, energy, and robotics.

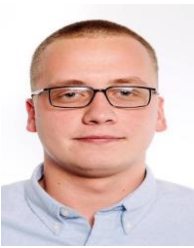

**VICTOR COBILEAN** received his B.Sc. degree in mechatronics from the Technical University of Cluj-Napoca, Romania, in 2020. He received his M. Sc. in Mechatronic Systems Engineering from Technical University of Cluj-Napoca, Romania, in 2022. He is currently working as a research assistant while reading for his doctoral degree in computer science at Virginia Commonwealth University, Richmond. His research interests include Anomaly Detection, Informed Machine Learning, Unsupervised Learning.

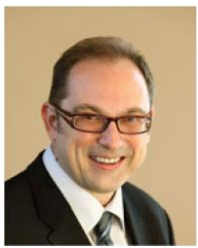

**MILOS MANIC** (SM'06-M'04-StM'96) is a Professor with the Computer Science Department and Director of VCU Cybersecurity Center at Virginia Commonwealth University, He completed over 40 research grants in AI/ML in cyber and energy and intelligent controls. He authored over 200 refereed articles, holds several U.S. patents, and has won the 2018 R\&D 100 Award for Autonomic Intelligent Cyber Sensor (AICS). He is a Fellow of IEEE, President-Elect of IEEE IES, Fellow of Commonwealth Cyber Initiative, Inductee of the National Academy of Inventors, and recipient of IEEE IES 2019 Anthony J.Hornfeck Service Award, 2012 J. David Irwin Early Career Award, 2017 IEM Best Paper Award, an associate editor of Transactions on Industrial Informatics, Open Journal of Industrial Electronics Society, IES Officer, and Senior AdCom member. He served as AE of Trans. on Industrial Electronics, was a founding chair of the IEEE IES Technical Committee on Resilience and Security in Industry, and was a general chair of IEEE IECON 2018, and IEEE HSI 2019.

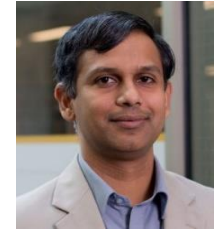

**JAYASIMHA ATULASIMHA** Jayasimha Atulasimha (SM'11) received the M.S. and Ph.D. degrees in aerospace engineering from the University of Maryland, College Park, MD, USA, in 2003 and 2006, respectively. He is a Professor of Mechanical and Nuclear Engineering with a courtesy appointment in Electrical and Computer Engineering with the Virginia Commonwealth University, Richmond, VA, USA where he is currently the Associate Director for the Institute of Sustainable Energy and Environment. His current research interests include nanomagnets/spintronics based nonvolatile memory, neuromorphic (brain-like) computing hardware and more recently quantum computing with spin Qubit. He is a fellow of the ASME.